\documentclass[runningheads]{llncs}

\usepackage{eccv}

\usepackage{eccvabbrv}

\usepackage{graphicx}
\usepackage{booktabs}
\usepackage{xcolor}

\def\eqref#1{(\ref{eq:#1})}

\def\xcomment#1{\textcolor[rgb]{.3,.3,.1}{\text{$/\!\!/$ {\em #1}}}}
\def\comment#1{\kern-1cm\xcomment{#1}}
\def\eqcomment#1{\kern-1cm\xcomment{#1}}

\usepackage{wrapfig}
\usepackage{algorithm}
\usepackage{algpseudocode}
\usepackage{amsmath}
\usepackage{multirow}
\usepackage{tabularx}
\usepackage{placeins}
\usepackage{float}

\usepackage{epsfig}
\usepackage{multirow}
\usepackage{makecell}
\usepackage{amsfonts}
\usepackage{enumitem}
\usepackage{array}
\usepackage{algorithm}
\usepackage{algpseudocode}
\usepackage[symbol]{footmisc}
\usepackage[accsupp]{axessibility}
\usepackage{blindtext}
\usepackage{graphicx}
\usepackage{mathtools}
\usepackage{arydshln}

\usepackage{tabularx}
\usepackage{adjustbox}
\usepackage{multirow}
\usepackage{booktabs}
\usepackage{enumitem}
\usepackage{float} 

\usepackage{colortbl}
\usepackage{amssymb}
\usepackage{pifont}
\newcommand{\cmark}{\ding{51}} 
\newcommand{\xmark}{\ding{55}} 

\algrenewcommand\algorithmicindent{0.75em}

\def\eqref#1{(\ref{eq:#1})}

\def\eqref#1{(\ref{eq:#1})}

\def\xcomment#1{\textcolor[rgb]{.3,.3,.1}{\text{$/\!\!/$ {\em #1}}}}
\def\comment#1{\kern-1cm\xcomment{#1}}
\def\eqcomment#1{\kern-1cm\xcomment{#1}}

\def\real{\mathbb{R}}

\def\vec{\operatorname{vec}}

\def\norm#1{\left\lVert#1\right\rVert}

\def\l2#1{\norm{#1}_2}

\usepackage[accsupp]{axessibility}  

\usepackage{hyperref}

\usepackage{orcidlink}

\begin{document}

\title{ARC-Loc: Leveraging Azimuthal Ray Convergence as a Geometric Cue for Direct Cross-View Localization} 

\titlerunning{ARC-Loc}

\author{Hyeongsik Kim$^{*}$\inst{1} \and
Mincheol Kim$^{*}$\inst{1} \and
Heejoon Moon\inst{2}\orcidlink{0000-0002-8325-6989} \and 
Je Hyeong Hong$^{\dagger}$\inst{1,2,3}\orcidlink{0000-0003-2797-553X}}

\authorrunning{H.Kim and M.Kim et al.}

\institute{Dept. of Artificial Intelligence Semiconductor Engineering, Hanyang University, Republic of Korea \and
Dept. of Artificial Intelligence, Hanyang University, Republic of Korea \and
Dept. of Electronic Engineering, Hanyang University, Republic of Korea \\
\email{khs06007,tlfj02,wilko97,jhh37@hanyang.ac.kr} \\
{\footnotesize $^{*}$Equal contribution. \quad $^{\dagger}$Corresponding author.}
\url{https://github.com/SpatialAILab/ARC-Loc}}

\maketitle
\vspace{-3mm}
\begin{abstract}
Cross-view localization (CVL) estimates the pose of a ground image by matching it to a geo-referenced satellite image. 
To bridge the extreme viewpoint gap, mainstream pipelines rely on Bird's-Eye-View (BEV) transformations or 2D-to-3D lifting. 
However, deriving 3D structures from a single ground image is fundamentally ill-posed, causing these methods to endure geometric distortions and computational costs during 3D lifting or BEV projection. Furthermore, relying on external depth foundation models to resolve this introduces latency and remains susceptible to noisy predictions.
In this work, we present a different approach inspired by a human navigation technique called \textit{resection}, that can perform direct ground-to-satellite image matching and localization without relying on external depth foundation models.
The key insights of our method are that (i) ground keypoints can be translated into azimuthal rays on the satellite map, and (ii) these rays ideally converge at the user location. 
Exploiting this geometric constraint through direct line-to-point correspondences, we introduce a minimal Azimuthal Ray Convergence (ARC) solver to identify the intersection alongside an ARC loss to optimize the matching network. 
By eliminating dependencies on computationally heavy BEV transformations and external depth foundation models, our approach achieves faster, memory-efficient inference, while its explicit feature matching ensures straightforward compatibility with existing frameworks.
Experiments on VIGOR and KITTI demonstrate that ARC-Loc maintains competitive localization accuracy compared to recent approaches, highlighting its practicality.
\end{abstract}

\section{Introduction}
\label{sec:intro}

Cross-view localization (CVL) aims to estimate the pose of a ground camera given a geo-referenced aerial image. This is particularly crucial for autonomous driving and urban planning in environments such as urban canyons, which are highly prone to GNSS denial and multipath interference~\cite{ben2011improving}. While consumer-grade IMUs can reliably provide stable global orientation~\cite{um7_datasheet, lsm6dsv_datasheet}, recovering the precise global position remains highly challenging. Existing HD map-based approaches that solve this demand massive costs for creation and maintenance~\cite{zang2024data}.
In response, CVL has emerged as a promising alternative by matching ground images against globally accessible, geo-referenced satellite imagery. 
Nevertheless, bridging the visual gap caused by the drastic viewpoint differences between the two domains remains a formidable task.

\begin{figure*}[t]
    \centering
    \includegraphics[width=1.0\linewidth]{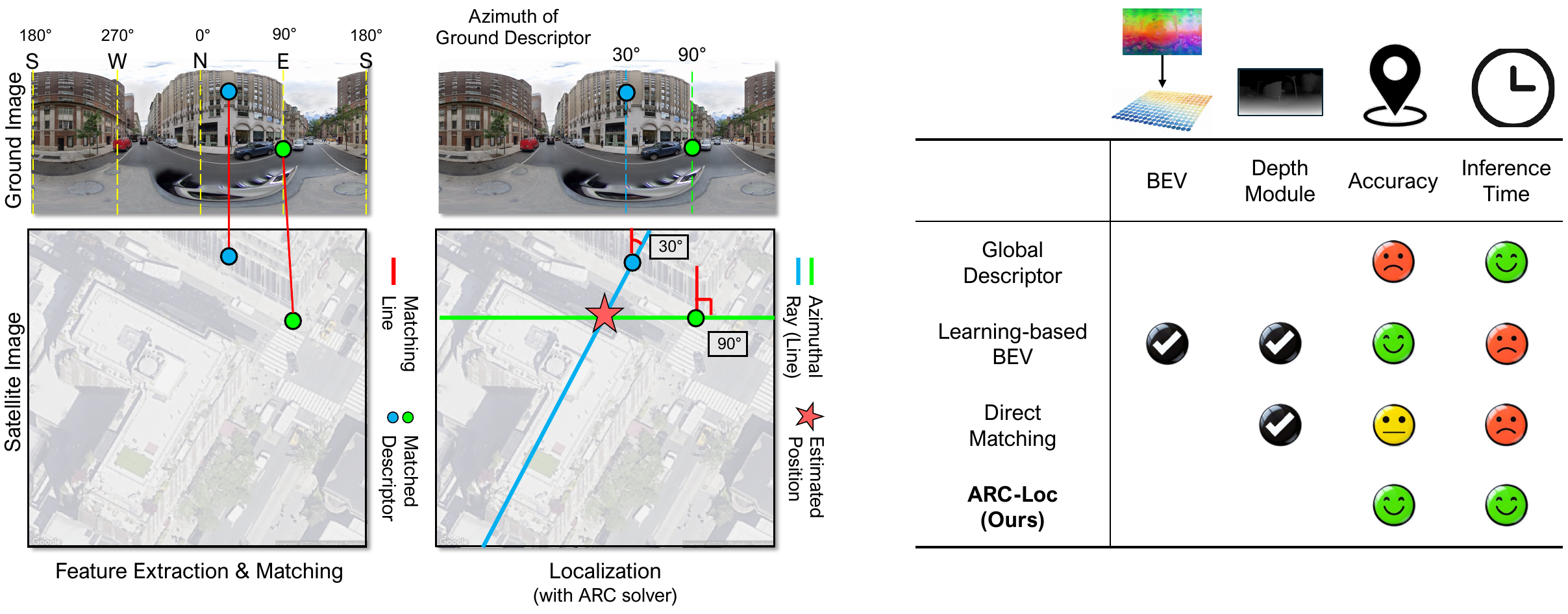}
    \vspace{-6mm}
    \caption{\textit{(Left)} \textbf{Localization procedure of ARC-Loc}, \textit{(Right)} \textbf{comparison of CVL approaches}. Given the matched descriptors across the ground image and the aerial satellite image, each ground keypoint’s azimuth is mapped into a line constraint on the satellite map by drawing an \emph{azimuthal ray} from the corresponding satellite keypoint along the ground-view azimuth. The user device must lie on this ray, and the location is obtained by the intersection point of these rays.
    Based on this mechanism, ARC-Loc does not require computationally heavy BEV transformations or external depth foundation models, achieving faster inference times and competitive localization accuracy.
    }
    \label{fig:overview}
    \vspace{-6mm}
\end{figure*}

As shown in~\cref{fig:overview}, prior CVL approaches generally fall into three categories, each facing distinct challenges in bridging the severe ground-to-aerial viewpoint gap. Global descriptor-based methods~\cite{slicematch, xia2022visual, ccvpe} estimate pose by comparing single-vector representations, inherently sacrificing the fine-grained geometric details needed for precise localization. To preserve spatial cues, recent methods project ground features into an orthographic Bird's-Eye-View (BEV) space~\cite{fg2, fervers2023uncertainty, bevsplat, 3dgs, li2024bevformer}. 
However, projecting 2D ground pixels into a BEV coordinate system fundamentally requires unavailable 3D priors, introducing geometric distortion and additional computation for estimating depth, either via implicit learning~\cite{fg2, fervers2023uncertainty, bevsplat} or by relying on depth foundation models~\cite{fg2,bevsplat}.

Furthermore, rendering a BEV view requires non-trivial height estimation to determine exactly which ground pixels correspond to surfaces actually visible from the satellite~\cite{fg2, orthogonal}, often leading to undesired geometric distortions. 
Alternatively, a direct matching-based framework such as Loc$^2$~\cite{loc2} bypasses BEV warping through direct 2D-2D correspondences, yet still relies on external depth priors for 3D lifting~\cite{unik3d}, introducing sensitivity to depth inaccuracies. 
Thus, existing paradigms leave ample room for cross-view localization strategies free from geometric distortions, computational cost, and/or external depth dependencies.

    A fundamental reason existing frameworks rely on complex 3D or BEV representations is their attempt to simultaneously resolve both translation and rotation. In the real-world applications such as autonomous vehicles~\cite{kitti, caesar2020nuscenes}, the reliable global orientation can be obtained by standard IMUs~\cite{um7_datasheet, lsm6dsv_datasheet} or orientation prediction networks~\cite{shi2024weakly,bevsplat}.
    Hence, we can reframe the cross-view localization task to focus on estimating precise position. 
    For this purpose, we draw inspiration from a traditional human navigation technique called \textit{resection}~\cite{resection}, in which an observer can determine his or her exact position by identifying visible landmarks, measuring the azimuth (bearing) to each landmark, and drawing corresponding rays on the 2D (aerial) map. The true location then lies at the convergence point of these azimuthal rays. We hypothesize that this purely 2D, line-intersection constraint can be directly adapted for cross-view localization, where ground-level visual features act as the azimuth measurements and the 2D aerial map serves as the global coordinate frame.
    
    Motivated by the above principle, we present \textbf{ARC-Loc}, a novel framework that leverages Azimuthal Ray Convergence (ARC) as a geometric cue for cross-view localization. Specifically, we can interpret direct 2D ground-to-2D satellite point correspondences as 2D line (azimuthal rays)-to-2D point constraints (\cref{fig:overview}), and subsequently the camera position can be estimated at the convergence point of these azimuthal rays.
    To implement this, we introduce the \textit{ARC solver}, a differentiable line-based solver capable of estimating the camera position from a minimal set of just two ground-aerial correspondences. Furthermore, to train the network for robust feature matching without dense correspondence labels, we propose the \textit{ARC loss}, which enforces both accurate position estimation and the convergence of the mapped azimuthal rays. 
    By operating entirely in the native 2D image domain without dependencies on external depth foundation models or BEV transformation, ARC-Loc avoids geometric distortions and achieves faster, memory-efficient inference.
    Moreover, ARC-Loc enables explicit, interpretable local feature matching similar to \cite{fg2,loc2}. This provides straightforward compatibility with geometric correspondence-based pipelines~\cite{fg2,loc2}, while seamlessly supporting RANSAC for robust cross-view localization.

\noindent To summarize, our key contributions are as follows:
\begin{itemize}
    \vspace{-1mm}
    \item\textbf{ARC-Loc:} a cross-view localization framework inspired by a traditional navigation technique that eliminates dependency on BEV transformation or external depth models, achieving faster and memory-efficient inference.
    \item\textbf{ARC solver:} a differentiable (closed-form) minimal solver for estimating camera position from minimum of two azimuthal ray-to-aerial point constraints that is also compatible with RANSAC for robust localization.
    \item\textbf{ARC loss:} a loss function based on the geometric constraint of azimuthal ray convergence for effective training of feature matching network without requiring dense correspondence labels or depth priors.
    \vspace{-2mm}
\end{itemize}
Extensive experiments on VIGOR and KITTI datasets demonstrate that ARC-Loc achieves competitive localization accuracy while eliminating reliance on BEV transformations and external depth foundation models, highlighting its practicability for real-world autonomous applications.

\section{Related work}
\label{sec:related_work}

\paragraph{\textbf{Global descriptor-based approaches.}}
Early cross-view localization largely treated the task as image retrieval by learning a shared embedding for ground and satellite views~\cite{vo2016localizing,cvmnet}.
Subsequent works showed that these global representations can be extended to metric localization by regressing pose from global features~\cite{zhu2021vigor}.
To improve spatial precision within this paradigm, later methods introduced finer matching mechanisms such as grid-structured descriptors~\cite{xia2022visual}, cyclic/rolling matching in CCVPE to address orientation uncertainty~\cite{ccvpe}, and cross-attention between ground features and satellite patches~\cite{cvtransloc,slicematch}.
Nevertheless, global-descriptor approaches inherently compress images into compact vectors, sacrificing geometric details, and their accuracy is also bounded by satellite patch resolution, making high-precision (\ie sub-meter) localization challenging.

\paragraph{\textbf{BEV-based approaches.}}
To bridge the extreme viewpoint gap between perspective ground views and orthographic satellite views, the dominant trend has shifted towards transforming ground images into an intermediate Bird's-Eye-View (BEV) representation. 
Initial approaches, such as DenseFlow~\cite{denseflow} and HC-Net~\cite{hc-net}, relied on explicit geometric projections like Inverse Perspective Mapping (IPM). 
While computationally efficient, IPM relies on the flat-world assumption, leading to geometric distortion or loss of vertical structures (\eg buildings, trees), which are critical landmarks in urban environments~\cite{bevsplat}.
To mitigate the geometric distortions, recent methods~\cite{boosting3dof,fervers2023uncertainty,fg2,li2024bevformer} employ learning-based transformations. 
GGCVT~\cite{boosting3dof} and Fevers~\etal.~\cite{fervers2023uncertainty} utilize the transformer model to learn the perspective-to-BEV mapping implicitly. 
More recently, FG$^2$~\cite{fg2} adapts the BEVFormer~\cite{li2024bevformer} architecture, lifting 2D image features into a 3D voxel space via deformable attention to synthesize a BEV feature map and performing 2D Procrustes alignment~\cite{procrustes} to estimate pose.
Similarly, BevSplat~\cite{bevsplat} leverages 3D Gaussian Splatting~\cite{3dgs} to render high-quality BEV features, where a depth foundation model~\cite{unik3d} is utilized to supervise the generation of 3D Gaussian primitives.
While recent learning-based BEV methods are elevating the localization accuracy, performing accurate BEV warping remains fundamentally challenging due to the inherent lack of exact 3D priors.
Furthermore, rendering a BEV view requires non-trivial height estimation to determine which ground pixels are actually visible from the satellite~\cite{bevsplat, orthogonal}.
Consequently, compensating for this missing 3D structure via explicit depth estimation or implicit BEV transformations~\cite{fg2, fervers2023uncertainty, bevsplat} not only demands heavy additional computation, but inherently leads to geometric distortions during BEV warping.

\paragraph{\textbf{Direct local feature matching-based approach.}}
The most recent work, Loc$^2$~\cite{loc2}, takes a slightly different approach by matching directly in the original domain.
They extract ground and aerial features from each respective image, lift the matched ground keypoints to the 3D space using a depth foundation model~\cite{unik3d}, and estimate pose via Procrustes alignment~\cite{procrustes}. 
Through this explicit matching approach, they provide greater interpretability compared to previous implicit approaches, enable the use of RANSAC for robust estimation, and mitigate the burden of BEV transformations.
Nevertheless, its reliance on an external depth foundation model not only introduces computational overhead but also makes the estimated pose highly sensitive to noisy depth predictions, resulting in slightly lower accuracy than recent CVL methods as shown in Tab.~\ref{tab:vigor_known}.

\section{Proposed method}
\label{sec:method}

We now illustrate the overall pipeline of our ARC-Loc method. 
Following several previous studies in cross-view localization~\cite{bevsplat,shi2024weakly}, our framework assumes the heading prior is provided to some degree of accuracy.
This is partly due to the fact that the ground heading can be readily obtained from consumer-grade IMUs~\cite{lsm6dsv_datasheet, um7_datasheet} (\eg modern electronic devices such as cellphones, cars, etc.) or orientation prediction network~\cite{shi2024weakly}.
Subsequently, our framework is primarily designed to estimate precise camera position under known orientation, although it remains robust to practical degree of orientation noise as shown in~\cref{fig:noise_analysis}. 

\begin{figure}[t]
    \centering
    \includegraphics[width=1.0\linewidth]{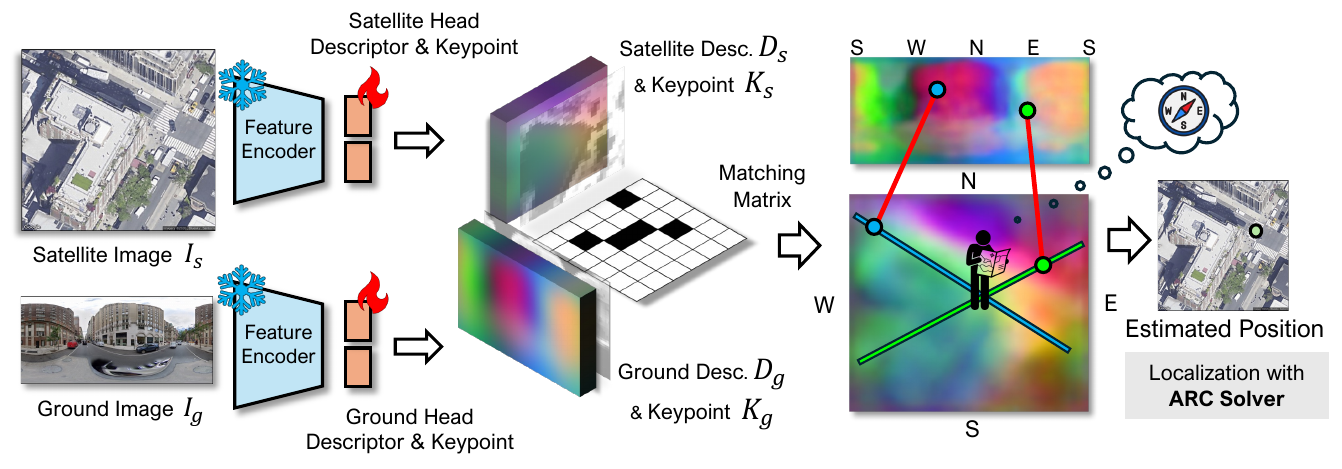}
    \caption{\textbf{Overview of ARC-Loc.}  The pipeline comprises two main stages: feature matching and localization. 
    Initially, we extract dense features using a pretrained visual feature encoder and pass them through keypoint and descriptor heads. 
    Then, by leveraging a matching matrix based on descriptor similarity and keypoint confidence, ARC-Loc establishes direct ground-aerial correspondences and transforms the ground matches into azimuthal rays. 
    ARC solver then estimates the camera position by finding the intersection of these rays. 
    During inference, as our solver requires a minimal set of just two correspondences, RANSAC can be equipped, enabling robust estimation.}
    \vspace{-4mm}
    \label{fig:main_architecture}
\end{figure}

\subsection{Framework overview}
\label{sec:overview}
The proposed framework processes a ground image $I_g$ and a geo-referenced satellite image $I_s$ to estimate the camera's position $\mathbf{p}\in\real^2$.
As shown in Fig.~\ref{fig:main_architecture}, the pipeline begins by extracting dense features to perform local feature matching, establishing direct 2D-2D cross-view correspondences between ground keypoints $\{\mathbf u_i \in \real^2\}$ and satellite keypoints $\{\mathbf x_i \in \real^2\}$. Each pair is then assigned with a matching confidence weight $w_i$, where $i$ denotes the correspondence index. 
Then, the matched ground keypoints are converted to respective azimuthal rays on the satellite image, providing geometric cues.
Finally, we estimate the camera position $\mathbf{p}$ using our proposed ARC solver.

\subsection{Cross-view feature extraction and direct matching}
\label{sec:feature_matching}
    
\paragraph{\textbf{Feature extraction.}} 
Following prior cross-view localization works that leverage vision foundation models~\cite{fg2,loc2,bevsplat}, we adopt the pretrained RADIOv3-L~\cite{ranzinger2024radio} as the backbone feature encoder.
Given the cross-view images $I_g$ and $I_s$, the backbone extracts dense feature maps $\mathbf{F}_g \in \mathbb{R}^{C\times H_g \times W_g}$ and $\mathbf{F}_s \in \mathbb{R}^{C \times H_s \times W_s}$, respectively, where $C$ denotes the output channel dimension.  
Since these models are known to provide rich semantic understanding and discriminative local representations, we anticipate these help to bridge the extreme visual gap between ground and satellite views.

The obtained features from the above encoder are then passed to two lightweight heads of identical architecture, namely a \textit{descriptor head} ($f_D$) and a \textit{keypoint head} ($f_K$) both of which extract necessary keypoint information for matching correspondences.
Both heads essentially consist of multiple convolutional layers followed by a self-attention layer to integrate contextual information.
Since we have separate heads for the ground view ($f_D^g$ and $f_K^g$) and satellite view ($f_D^s$ and $f_K^s$), we yield two dense descriptor maps ($\mathbf D_g$, $\mathbf D_s$) and two keypoint confidence maps ($\mathbf K_g$, $\mathbf K_s$) defined as follows:

\begin{equation}
    \vspace{-1mm}
    \begin{aligned}
        \mathbf{D}_g &= f_D^g(\mathbf{F}_g) \in \mathbb{R}^{\frac{C}{8} \times H_g \times W_g}, & \mathbf{D}_s &= f_D^s(\mathbf{F}_s) \in \mathbb{R}^{\frac{C}{8} \times H_s \times W_s}, \\
        \mathbf{K}_g &= f_K^g(\mathbf{F}_g) \in \mathbb{R}^{H_g \times W_g}, & \mathbf{K}_s &= f_K^s(\mathbf{F}_s) \in \mathbb{R}^{H_s \times W_s}.
    \end{aligned}
    \vspace{-1mm}
    \label{eq:feature_extraction}
\end{equation}

\paragraph{\textbf{Direct feature matching.}}
To compute the matching probabilities between cross-view image descriptors, we first define the pairwise similarity score as $Sim_{mn} = \text{sim}(\mathbf{d}_g^m, \mathbf{d}_s^n) / \tau$, where $\text{sim}(\cdot, \cdot)$ denotes the cosine similarity.
Here, $\mathbf{d}_g^m$ and $\mathbf{d}_s^n$ denote the $m$-th and $n$-th descriptors within the flattened maps $\mathbf{D}_g$ and $\mathbf{D}_s$, respectively.  The indices $m \in \{1, \dots, H_g W_g\}$ and $n \in \{1, \dots, H_s W_s\}$ represent the 1D spatial positions, and $\tau$ is a temperature hyperparameter.
Since not all features are matchable, we employ a dual-softmax operator with a learnable dustbin parameter $z$. The matching probability $\mathbf{M}_{mn}$ is then computed as:

\begin{equation}
\mathbf{M}_{mn} = \frac{\exp(Sim_{mn})}{\sum_{k=1}^{H_s W_s} \exp(Sim_{mk}) + \exp(z)} \cdot \frac{\exp(Sim_{mn})}{\sum_{l=1}^{H_g W_g} \exp(Sim_{ln}) + \exp(z)}.
\label{eq:dual_softmax}
\end{equation}
Inspired by~\cite{mickey}, we then compute the final matching confidence weight matrix $\mathbf{W}_{mn} = \mathbf{M}_{mn} \cdot \mathbf{k}_g^m \cdot \mathbf{k}_s^n,$
where $\mathbf{k}_g^m$ and $\mathbf{k}_s^n$ denote the keypoints' confidence scores at indices $m$ and $n$ from the flattened maps $\mathbf{K}_g$ and $\mathbf{K}_s$, respectively.
This integrates descriptor matching probability with keypoint selection confidence, enabling the matching module to favor correspondences supported by both strong descriptor similarity and reliable keypoint predictions.
Based on $ \{ \mathbf{W}_{mn} \}$, we select the top-$N$ correspondences with the highest similarity scores across all possible combinations of ground and satellite keypoints.
Consequently, the $i$-th selected correspondence ($i \in \{1, \dots, N\}$) consists of the satellite 2D keypoint $\mathbf{x}_i$, corresponding ground keypoint $\mathbf{u}_i$ and the matching confidence weight $w_i$.

\subsection{Azimuthal ray generation}
\label{sec:ray_gen}
For each 2D ground-to-2D satellite keypoint correspondence, the respective azimuthal ray $\mathbf l_i \in \mathbb{P}^2$ is constructed by drawing a line on the satellite map crossing the satellite keypoint ($\mathbf{x}_i$) along the 2D direction vector $\mathbf{r}(\phi_i) \in\mathbb{R}^2$, which has a slope of $\tan(\pi/2 - \phi_i)$ with respect to the global $x$-axis of the North-up satellite map.
This yields a 2D point-to-line geometric constraint between the aerial keypoint and the azimuthal ray, expressed as
\begin{equation}
\mathbf{l}_i = \mathbf{x}_i + \lambda \mathbf{r}(\phi_i), \quad \mathbf{l}_i^\top\mathbf{x}_i= 0,
\label{eq:azi_line}
\end{equation}
where $\lambda \in \mathbb{R}$ denotes a scaling parameter that spans the entire infinite line. 
As the camera location $\mathbf{p}$ ideally exists at the intersection of all azimuthal rays $\{ \mathbf{l}_i \}_{i=1}^N$, we can formulate as $\mathbf{p} =\bigcap_{i=1}^N \mathbf{l}_i$.

\subsection{Localization with the ARC solver}
\label{sec:pose_estimation}

After obtaining the 2D ground line-to-2D aerial point correspondences and their geometric constraints, ARC-Loc estimates the camera position through a closed-form solution derived in~\cref{eq:closed_form}.
To this end, we introduce the geometric formulation of our ARC minimal solver, followed by robust estimation.
Then, we describe a detailed strategy to handle noisy orientation priors in real-world cases.

\paragraph{\textbf{ARC solver.}} 
As mentioned in ~\cref{sec:ray_gen}, the ground camera ideally lies exactly at the intersection of the mapped azimuthal rays.
In practice, however, correspondences inevitably contain false-positive matches that prevent the strict ray convergence at single point. 
To address this, our ARC solver relaxes the strict intersection constraint into a weighted line-to-point distance minimization problem, where the confidence weight $w_i$ implicitly prioritizes reliable correspondences during optimization.
Thus, the optimal camera position $\hat{\mathbf{p}}$ is derived as:
\begin{equation}
    \hat{\mathbf{p}} = \underset{\mathbf{p}}{\text{argmin}} \sum_{i=1}^{N} w_i \cdot \text{dist}(\mathbf{p}, \mathbf{l}_i)^2,
    \label{eq:obj_function}
\end{equation}
where $\text{dist}(\cdot, \cdot)$ and $N$ denote the perpendicular line-to-point distance and number of correspondences, respectively. 
To effectively solve the objective function in \cref{eq:obj_function}, we cast the minimization as a weighted linear least-squares problem.
We first define the unit normal vector of the $i$-th azimuthal line as $\mathbf{n}_i = [\sin(\phi_i), \cos(\phi_i)]^\top$.
Then, the perpendicular distance from the candidate position $\mathbf{p}$ to the $i$-th azimuthal ray ($\mathbf{l}_i$) can be calculated by projecting the vector $(\mathbf{p} - \mathbf{x}_i)$ onto the line's normal vector $\mathbf{n}_i$, which is expressed as $\mathbf{n}_i^\top (\mathbf{p} - \mathbf{x}_i)$.
By substituting this into \cref{eq:obj_function}, we can derive the optimal position $\hat{\mathbf{p}}$ in a closed form as
\begin{equation}
    \hat{\mathbf{p}} = (\mathbf{A}^\top \boldsymbol{\mathrm{\Omega}} \mathbf{A})^{-1} \mathbf{A}^\top \boldsymbol{\mathrm{\Omega}} \mathbf{b},
    \label{eq:closed_form}
\end{equation}
$\mathbf{A} \in \mathbb{R}^{N \times 2}$ is the design matrix with its $i$-th row as $\mathbf{n}_i^\top$, $\mathbf{b} \in \mathbb{R}^{N}$ is the target vector with $b_i = \mathbf{n}_i^\top \mathbf{x}_i$, and $\boldsymbol{\mathrm{\Omega}} = \operatorname{diag}(w_1, \dots, w_N)$ is the diagonal weight matrix.

Note, \cref{eq:closed_form} implies two key aspects. 
First, $\hat{\mathbf{p}}$ can be estimated using two minimal correspondences. 
Second, since $\hat{\mathbf{p}}$ is a closed-form solution of $\{w_i\}$, this can be efficiently used for end-to-end training of our pipeline via pose supervision.  
The full derivation of \cref{eq:closed_form} is provided in the supplement~\cite{supmat}.

\paragraph{\textbf{Robust estimation.}}
During the inference, as cross-view matching inherently produces a large number of false-positive matches, we equip the ARC-solver with the RANSAC~\cite{ransac} for robust position estimation. 
To select the best position hypothesis $\mathbf{p}_{\mathrm{hyp}}$, we use a line-to-point distance-based score function defined as:
\begin{equation}
\mathcal{S}(\mathbf{p}_{\mathrm{hyp}}) \;=\; \sum_{i=1}^{N} \frac{w_i}{1 + \left(\text{dist}(\mathbf{p}_{\mathrm{hyp}}, \textbf{l}_i)/\sigma\right)^2},
\label{hyp_score}
\end{equation}
where $\sigma$ is a scaling parameter and $\textbf{l}_i$ denotes the $i$-th azimuthal line.
During inference, as the ARC solver requires a minimal two line-point correspondences to generate $\mathbf{p}_{\mathrm{hyp}}$, we sample multiple hypotheses and select the one maximizing $\mathcal{S}(\mathbf{p}_{\mathrm{hyp}})$ as the initial position. 
The final position is then obtained via~\cref{eq:closed_form} using all inlier correspondences with the initial position.

\paragraph{\textbf{Extension for handling noisy orientation.}}
While the global orientation can be obtained from consumer-grade IMUs or orientation prediction networks, these initial estimates inevitably contain some degree of noise~\cite{lsm6dsv_datasheet}.
To handle such perturbed conditions, we generate a set of fine-grained heading candidates ($\{\phi_{hyp}\}$) by dividing the search range $[\phi_0-\Delta,\ \phi_0+\Delta]$ into equally spaced intervals, where $\phi_0$ denotes the noisy orientation prior, and $\Delta$ denotes the expected deviation of the orientation noise.
For each candidate, we update the sets of azimuthal rays {$\{ \mathbf{l}_i \}_{i=1}^N$} in \cref{eq:azi_line} with candidate orientation ($\phi_{hyp}$). 
Then, for each candidate, we estimate the corresponding position ($\mathbf{p}_{\mathrm{hyp}}$) with RANSAC using the quality score in \cref{hyp_score}.
Consequently, this process yields a set of paired hypotheses $(\phi_{\mathrm{hyp}}, \mathbf{p}_{\mathrm{hyp}})$.
We then select the best orientation candidate that maximizes \cref{hyp_score} and use its associated camera position as the final estimation.

Although executing RANSAC for multiple heading candidates may seem to incur high computational cost at first instance, we can efficiently mitigate this overhead at the implementation level by leveraging a parallelized RANSAC framework on modern GPUs such as NVIDIA A100.

\subsection{Loss function}
\label{sec:loss}
\begin{figure}[t]
    \centering
    \includegraphics[width=1.0\linewidth]{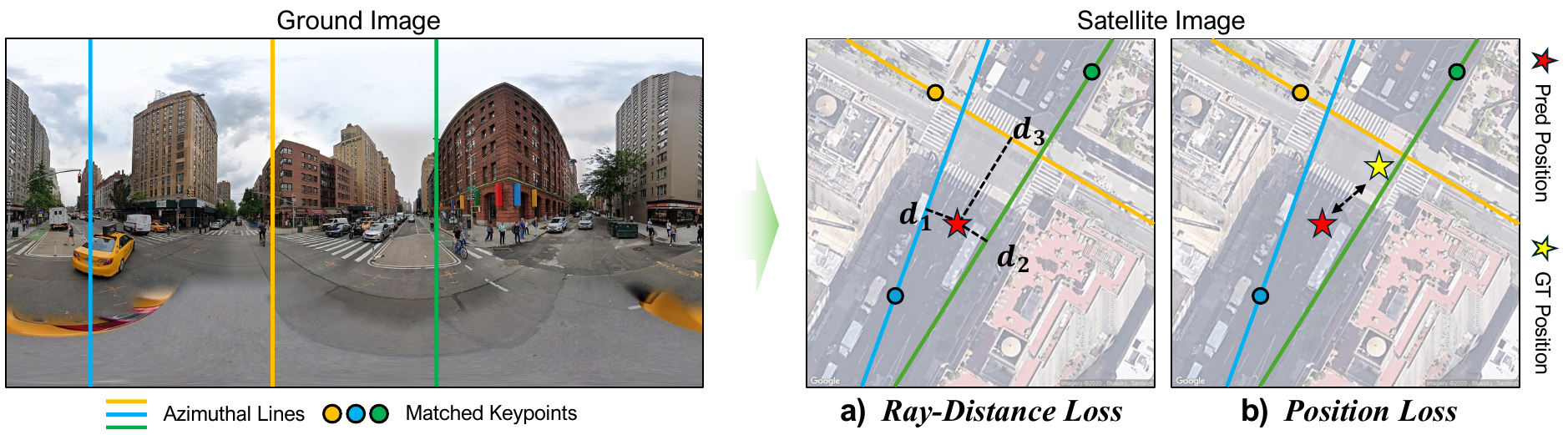}
    \caption{
    \textbf{ARC loss visualization.} Our ARC loss aims to supervise the ARC-Loc framework in an end-to-end manner. (a) The ray-distance loss enforces the azimuthal lines to converge at the predicted location without requiring explicit matching labels, while (b) the position loss directly minimizes the discrepancy between the predicted and ground-truth camera positions. By backpropagating these geometric errors, the network implicitly learns feature matching by penalizing geometrically inconsistent outliers.
    }
    \label{fig:self_consistency_loss}
    \vspace{-0.5cm}
\end{figure}

To train the ARC-Loc framework in an end-to-end manner, we introduce the Azimuthal Ray Convergence (ARC) loss (\cref{fig:self_consistency_loss}) $\mathcal{L}_\text{ARC}$, which is a composite of the position loss ($\mathcal{L}_{position}$) and the (azimuthal) ray-distance loss ($\mathcal{L}_{ray-dist}$).
Specifically, $\mathcal{L_\text{ARC}} = \mathcal{L}_{position} + \alpha \mathcal{L}_{ray-dist}$, where $\alpha$ is a hyperparameter to balance the two terms.
$\mathcal{L}_{position}$ is defined as $\| \mathbf{p}_{gt} - \hat{\mathbf{p}} \|_2^2$, where $\mathbf{p}_{gt}$ is the ground-truth position and $\hat{\mathbf{p}}$ is the estimated position, providing direct supervision signal for the camera location.
On the other hand, $\mathcal{L}_{ray-dist}$ is defined as $\sum_{i=1}^{N} \bar{w}_i \cdot \text{dist}(\hat{\mathbf{p}}, \textbf{l}_i)^2$, where $\{\bar{w}_i\}$ represent a set of normalized weights obtained by applying the softmax operation to the matching scores $\{ w_i \}$.
By leveraging the solver's predicted position $\hat{\mathbf{p}}$ as a geometric anchor, this objective explicitly penalizes rays that fail to converge. Since this loss is weighted by the matching confidence $\bar{w}_i$, the keypoint and descriptor heads (\cref{sec:feature_matching}) learn to lower the matching scores of correspondences yielding geometrically inconsistent azimuthal rays, acting as an effective geometric regularization technique for cross-view localization.

\section{Experiments}
\subsection{Datasets and evaluation metrics}

\paragraph{\textbf{VIGOR dataset.}} VIGOR~\cite{zhu2021vigor} is a benchmark for fine-grained cross-view localization, comprising $360^\circ$ ground panoramas and $70~\text{m} \times 70~\text{m}$ geo-referenced satellite images from four cities (Chicago, NewYork, SanFrancisco, Seattle). 
It provides Same-Area (unseen locations) and Cross-Area (unseen cities) settings for evaluation. To assess orientation robustness, we use both orientation-aligned panoramas and those with noisy orientations applied via horizontal cyclic shifts.

\paragraph{\textbf{KITTI dataset.}} KITTI~\cite{kitti, beyond, cvlnet} is a widely used benchmark featuring perspective images from front-facing cameras paired with $100~\text{m} \times 100~\text{m}$ geo-referenced satellite maps.
An orientation prior is typically provided with an added noise level of approximately $\pm10^\circ$ to simulate the inaccuracies of consumer-grade IMU/compass. 
The dataset is partitioned into Training, Test1, and Test2 sets. 
While the Training and Test1(Same-Area) sets consist of different measurements taken from the same region in the training data sequence, the Test2(Cross-Area) set contains data from unseen regions.

\paragraph{\textbf{Metrics.}} We evaluate ARC-Loc using mean and median position and orientation errors on both datasets.
For KITTI, we additionally report recall at position ($R@1m, R@5m$) and orientation ($R@1^\circ, R@5^\circ$) thresholds, and decompose the position error into longitudinal and lateral components.
We further assess localization accuracy given known orientations as well as under noisy prior conditions designed to reflect the practical error ranges of consumer-grade IMUs.
Finally, we report inference latency and memory usage to evaluate practical deployability.

\paragraph{\textbf{Implementation details.}}
We employ pretrained RADIOv3-L~\cite{ranzinger2024radio} as our feature backbone, keeping weights frozen to preserve generalized representations. 
Input images are resized to $720 \times 1440$ for ground panoramas and $656 \times 656$ for satellite images. 
Both descriptor and keypoint heads project features into a 128-D space. 
During matching, we compute cosine similarity with a temperature of $\tau=0.1$ and select the top $N=512$ correspondence pairs to generate azimuthal rays. 
The model is trained using the Adam optimizer with the initial learning rate of $10^{-4}$ and a Cosine Annealing scheduler. 
We set the ARC loss coefficient $\alpha$ to 0.5 and train with a batch size of 64 on a single NVIDIA A100 GPU. 
For the KITTI dataset, we adopt the orientation estimator from~\cite{shi2024weakly}.

\begin{table}[t]
\centering
\caption{Localization results on the VIGOR dataset under the known orientation setting. The best and second-best results within each category (BEV and Non-BEV) are highlighted in \textbf{bold} and \underline{underlined}, respectively.}
\scriptsize
\renewcommand{\arraystretch}{1}

\begin{tabular*}{\linewidth}{@{\extracolsep{\fill}}llcccc@{}}
\toprule
\multirow{3.5}{*}{\textbf{Type}} & \multirow{3.5}{*}{\textbf{Models}} & \multicolumn{2}{c}{\textbf{Same-Area}} & \multicolumn{2}{c}{\textbf{Cross-Area}} \\
\cmidrule(lr){3-4} \cmidrule(lr){5-6}
& & \multicolumn{2}{c}{$\downarrow$ Localization (m)} & \multicolumn{2}{c}{$\downarrow$ Localization (m)} \\
& & Mean & Median & Mean & Median \\
\midrule

\multirow{6}{*}{BEV} 
 & GGCVT & 4.12 & 1.34 & 5.16 & 1.40 \\
 & DenseFlow & 3.03 & \textbf{0.97} & 5.01 & 2.42 \\
 & G2SWeakly & 4.19 & 1.68 & 4.70 & 1.68 \\
 & HC-Net & \underline{2.65} & 1.17 & 3.35 & 1.59 \\
 & FG$^2$ & \textbf{1.95} & \underline{1.08} & \textbf{2.41} & \underline{1.37} \\
 & BevSplat & 2.87 & 1.58 & \underline{2.84} & \textbf{1.36} \\ 

\cmidrule(lr){1-6} 

\multirow{4}{*}{Non-BEV} 
 & SliceMatch & 5.18 & 2.58 & 5.53 & 2.55 \\
 & CCVPE & 3.60 & \underline{1.36} & 4.97 & \underline{1.68} \\
 & Loc$^2$ & \underline{3.06} & 1.59 & \underline{3.43} & 1.90 \\
 & ARC-Loc (ours) & \textbf{2.52} & \textbf{1.20} & \textbf{3.11} & \textbf{1.59} \\ 

\bottomrule
\end{tabular*}
\label{tab:vigor_known}
\end{table}

\begin{figure}[t]
    \centering
    \begin{minipage}[c]{0.55\textwidth}
        \centering
        \includegraphics[width=\textwidth]{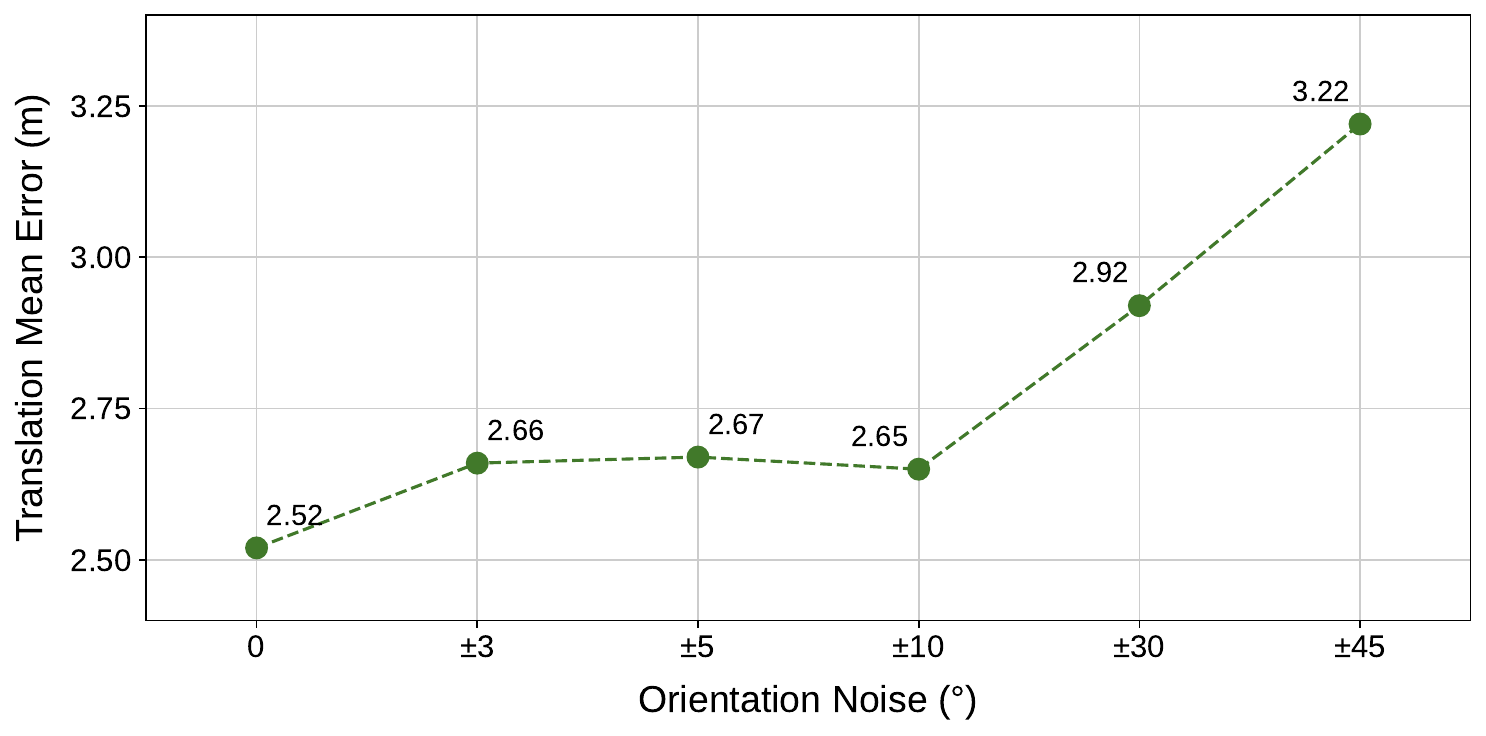} 
        \vspace{-6.0mm}
        \caption{\textbf{Impact of orientation noise.} Mean position error under bounded heading uncertainty ($\pm\Delta^\circ$), accounting for IMU errors ($\pm5.9^\circ$)~\cite{lsm6dsv_datasheet} and environmental disturbances.}
        \label{fig:noise_analysis}
    \end{minipage}
    \hfill
    \begin{minipage}[c]{0.43\textwidth}
        \vspace{-1mm}
        \centering
        \includegraphics[width=\textwidth]{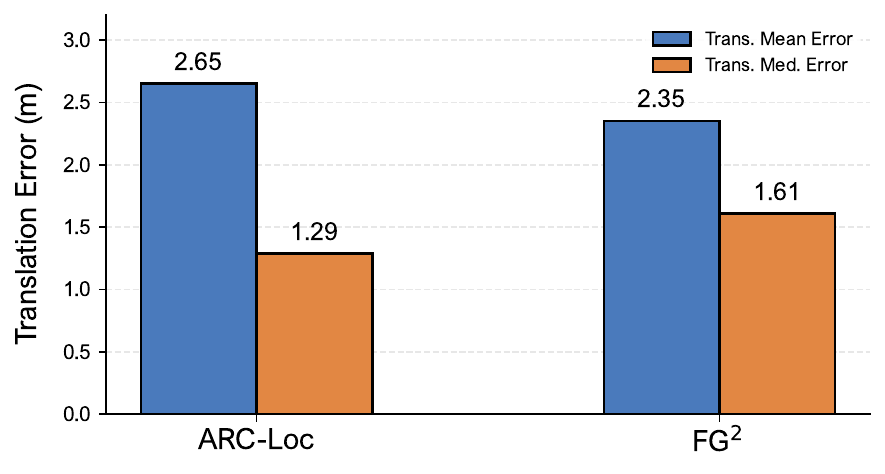} 
        \caption{Position mean and median errors of ARC-Loc versus $FG^2$ given $\pm10^\circ$ orientation noise.}
        \label{fig:orientation_vs_fg2}
    \end{minipage}
    \vspace{-2mm}
\end{figure}

\begin{table}[t]
\centering
\caption{Quantitative comparison on the KITTI dataset under orientation noise ($\pm10^\circ$). The best and second-best results within each category (BEV and Non-BEV) are highlighted in \textbf{bold} and \underline{underlined}, respectively. * denotes the use of the orientation estimator from G2SWeakly~\cite{shi2024weakly}.}
\label{tab:kitti_stacked}
\vspace{-2mm}
\resizebox{\linewidth}{!}{
\small 
\setlength{\tabcolsep}{4pt}
\begin{tabular}{llcccccccccc}
\toprule
\multirow{2}{*}{Type} & \multirow{2}{*}{Models} & \multicolumn{2}{c}{Loc. (m) $\downarrow$} & \multicolumn{2}{c}{Lateral (\%)$\uparrow$} & \multicolumn{2}{c}{Long. (\%)$\uparrow$} & \multicolumn{2}{c}{Ori. ($^\circ$) $\downarrow$} & \multicolumn{2}{c}{Ori. $\uparrow$} \\
\cmidrule(lr){3-4} \cmidrule(lr){5-6} \cmidrule(lr){7-8} \cmidrule(lr){9-10} \cmidrule(lr){11-12}
 & & Mean & Median & R@1m & R@5m & R@1m & R@5m & Mean & Median & R@$1^\circ$ & R@$5^\circ$ \\

\midrule

\multicolumn{12}{c}{\textbf{Same-Area}} \\ \midrule

\multirow{5}{*}{BEV} 
   & GGCVT & - & - & 76.44 & 98.89 & 23.54 & 62.18 & - & - & \underline{99.10} & \textbf{100.0}\\
   & DenseFlow & 1.48 & \textbf{0.47} & 95.47 & \textbf{99.79} & 87.89 & 94.78 & 0.49 & \underline{0.30} & 89.40 & 99.31 \\
   & HC-Net & \underline{0.80} & \underline{0.50} & \textbf{99.01} & \underline{99.73} & \underline{92.20} & \textbf{99.25} & \underline{0.45} & 0.33 & 91.35 & \underline{99.84}\\
   & FG$^2$ & \textbf{0.75} & 0.51 & \underline{95.81} & 99.66 & \textbf{92.50} & \underline{99.05} & 0.93 & 0.66 & 67.27 & 98.91\\
   & BevSplat* & 2.87 & 2.06 & 60.28 & 94.24 & 35.62 & 76.57 & \textbf{0.33} & \textbf{0.28} & \textbf{99.99} & \textbf{100.0}\\
 \cmidrule{1-12}
\multirow{3}{*}{\shortstack{Non-\\BEV}} 
& CCVPE & 1.22 & \underline{0.62} & \underline{97.35} & \underline{99.71} & \underline{77.13} & 97.16 & \underline{0.67} & \underline{0.54} & \underline{77.39} & \underline{99.95} \\
& Loc$^2$ & \underline{1.13} & 0.77 & 93.59 & \textbf{99.97} & 71.51 & \underline{98.17} & 1.97 & 1.43 & 36.68 & 92.84 \\
& ARC-Loc* (ours)& \textbf{0.79} & \textbf{0.50} & \textbf{97.72} & 99.63 & \textbf{87.01} & \textbf{98.60} & \textbf{0.21} & \textbf{0.18} & \textbf{99.87} & \textbf{100.0} \\

\midrule
\multicolumn{12}{c}{\textbf{Cross-Area}} \\ \midrule

\multirow{5}{*}{BEV} 
   & GGCVT & - & - & 57.72 & 91.16 & 14.15 & 45.00 & - & - & \underline{98.98} & \textbf{100.0}\\
   & DenseFlow & 7.97 & \underline{3.52} & 54.19 & 91.74 & 23.10 & \underline{61.75} & \underline{2.17} & \underline{1.21} & 43.44 & \underline{89.31}\\
   & HC-Net & 8.47 & 4.57 & \textbf{75.00} & \textbf{97.76} & \textbf{58.93} & \textbf{76.46} & 3.22 & 1.63 & 33.58 & 83.78\\
   & FG$^2$ & \underline{7.31} & 4.15 & 37.89 & 85.65 & 21.98 & 60.77 & 3.62 & 2.37 & 23.03 & 77.84 \\
   & BevSplat* & \textbf{6.20} & \textbf{2.51} & \underline{60.01} & \underline{95.17} & \underline{27.41} & 60.45 & \textbf{0.33} & \textbf{0.28} & \textbf{99.99} & \textbf{100.0}\\
 \cmidrule{1-12}
\multirow{3}{*}{\shortstack{Non-\\BEV}} 
   & CCVPE & 9.16 & \underline{3.33} & \underline{44.06} & \underline{90.23} & \underline{23.08} & \underline{64.31} & \underline{1.55} & \underline{0.84} & \underline{57.72} & \underline{96.19}\\
   & Loc$^2$ & \textbf{5.60} & \textbf{3.01} & \textbf{45.29} & \textbf{92.43} & \textbf{27.01} & \textbf{68.26} & 3.32 & 2.12 & 26.03 & 80.68\\
   & ARC-Loc* (ours) & \underline{7.45} & 4.32 & 38.45 & 87.13 & 20.26 & 59.85 & \textbf{0.20 }& \textbf{0.18} & \textbf{99.93} & \textbf{100.0} \\
\bottomrule
\end{tabular}}
\label{tab:kitti}
\vspace{-5mm}
\end{table}

\subsection{Localization results}
To demonstrate the effectiveness of our proposed ARC-Loc, we compare ARC-Loc against the state-of-the-art cross-view localization methods, including BEV-based approaches~\cite{fg2, bevsplat, hc-net, denseflow, boosting3dof, shi2024weakly} and Non-BEV approaches~\cite{loc2, ccvpe, slicematch}.

\begin{figure}[!t]
    \centering
    \includegraphics[width=1.0\linewidth]{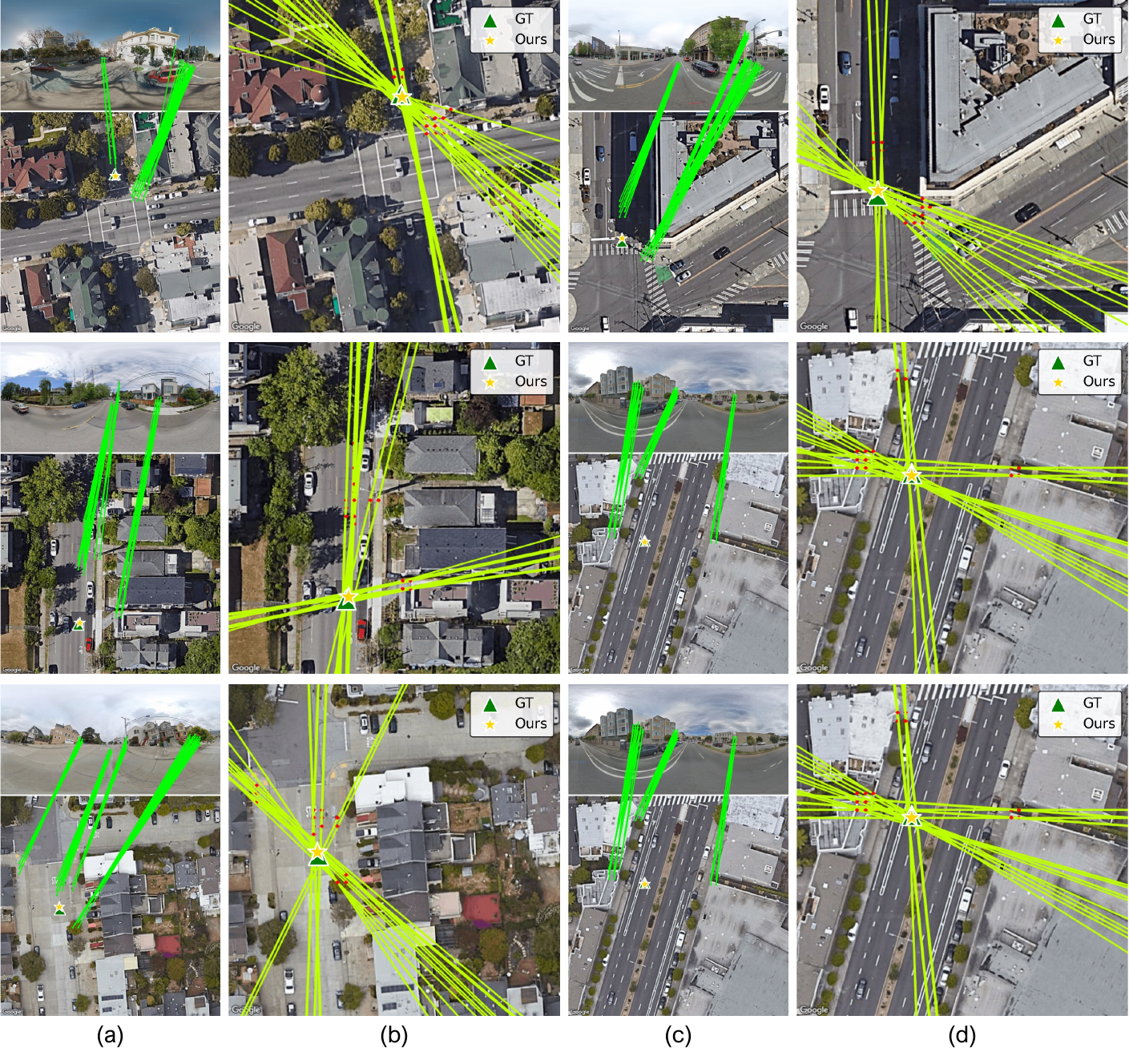}
    \vspace{-6mm}
    \caption{Visualization of feature matching and azimuthal rays. (a), (c) denote cross-view feature matching, and (b), (d) denote azimuthal rays for corresponding matches.}    
    \label{fig:qualitative_results}
    \vspace{-6mm}
\end{figure}

\paragraph{\textbf{VIGOR.}}  
In \cref{tab:vigor_known}, we first evaluate ARC-Loc under the known orientation setting. 
Notably, our method achieves the lowest position error among all Non-BEV methods. Specifically, it outperforms Loc$^2$ without relying on any external depth modules, reducing the mean localization error by 17.6\% on the Same-Area and by 9.3\% on the Cross-Area.
Compared to BEV-based methods, ARC-Loc yields a slightly higher error than the state-of-the-art $FG^2$~\cite{fg2}, while still exhibiting lower mean position errors than several BEV-based approaches, demonstrating its highly competitive localization accuracy.
In Overall, ARC-Loc achieves the second-best overall mean error on the Same-Area setting, closely following FG$^2$.
We visually confirm this mechanism in \cref{fig:qualitative_results}, under the known orientation setting, the extracted azimuthal rays converge around the ground-truth location, highlighting the interpretability of our framework without explicit 3D lifting.

While the previous results validate the effectiveness of our position-focused design under known orientation, real-world deployment often involves noisy sensor data. 
To evaluate our framework's robustness in such practical scenarios, we further evaluate ARC-Loc under bounded heading uncertainties. \cref{fig:noise_analysis} demonstrates our model's stability, showing that the mean position error remains within 2.65 m for noise levels up to $\pm10^\circ$, before gradually increasing at larger noise bounds (e.g., $\pm45^\circ$). 
Notably, despite having a higher mean position error than FG$^2$ under the practical $\pm10^\circ$ noisy orientation setting (\cref{fig:orientation_vs_fg2}), ARC-Loc achieves a lower median position error of 1.29 m.
This confirms that our azimuthal ray based approach is effective even under orientation noise.

\paragraph{\textbf{KITTI.}}
We evaluate ARC-Loc against both BEV-based and Non-BEV methods. 
As shown in \cref{tab:kitti}, ARC-Loc demonstrates competitive position accuracy on the Same-Area setting, performing on par with leading BEV-based approaches like FG$^2$ and outperforming existing Non-BEV baselines. 
On the Cross-Area setting, our method yields a higher position error than Loc$^2$, but it maintains stable performance. 
This result can be attributed to our line-based geometric formulation, which relies on precise azimuthal cues that are inherently more challenging to extract under the restricted FoV and unseen layouts of the Cross-Area setting. 
Nevertheless, these findings suggest that ARC-Loc effectively balances computational efficiency with localization accuracy, positioning it as a feasible alternative to more computationally intensive models.

\paragraph{\textbf{Inference and memory efficiency.}}
To evaluate practicality on resource-constrained platforms, we compare inference latency and memory consumption on VIGOR across inference configurations (\cref{tab:efficiency_comparison}).
Here, matching-stage latency includes BEV feature construction or depth inference when required.
Since ARC-Loc performs direct matching in the native 2D image domain without BEV transformation or depth foundation modules, it achieves substantially lower matching latency while maintaining a compact memory footprint compared to depth-dependent methods such as Loc$^2$ and BevSplat. 
These results show that ARC-Loc improves efficiency by removing the overhead of BEV construction and depth-based 3D lifting, while maintaining competitive localization accuracy.

\vspace{-4mm}
\begin{table}[!h]
\centering
\caption{Inference latency and memory comparison with ViT-based models. Matching Latency and Pose Est. Latency denote the feature matching and solver/RANSAC stages, respectively. Original denotes each method's official sequential RANSAC implementation, while Parallel (ours) denotes our GPU-parallel RANSAC implementation.}
\label{tab:efficiency_comparison}
\vspace{-2mm}
\scriptsize
\setlength{\tabcolsep}{2.5pt}
\renewcommand{\arraystretch}{0.95}
\resizebox{\linewidth}{!}{%
\begin{tabular}{lcccccc}
\toprule
\textbf{Method} 
& \textbf{BEV} 
& \makecell{\textbf{Depth}\\\textbf{Module}} 
& \textbf{RANSAC} 
& \makecell{\textbf{Matching} $\downarrow$\\\textbf{Latency (ms)}} 
& \makecell{\textbf{Pose Est.} $\downarrow$\\\textbf{Latency (ms)}} 
& \makecell{\textbf{Memory}\\\textbf{(MB)} $\downarrow$} \\
\midrule

\multirow{3}{*}{FG$^2$} 
& \multirow{3}{*}{\cmark} 
& \multirow{3}{*}{\xmark} 
& \xmark 
& 187.61 & 2.03 & 885 \\
& & & Original 
& 185.94 & 3630 & 910 \\
& & & \makecell{Parallel (ours)} 
& 187.34 & 39.67 & 1281 \\

\midrule
BevSplat 
& \cmark 
& \cmark 
& -- 
& \multicolumn{2}{c}{139.90 (end-to-end)} 
& 2577 \\

\midrule
\multirow{3}{*}{Loc$^2$} 
& \multirow{3}{*}{\xmark} 
& \multirow{3}{*}{\cmark} 
& \xmark 
& 137.98 & 1.44 & 2131 \\
& & & Original 
& 136.13 & 3984 & 2168 \\
& & & \makecell{Parallel (ours)} 
& 137.52 & 28.32 & 2810 \\

\midrule
\multirow{2}{*}{ARC-Loc} 
& \multirow{2}{*}{\xmark} 
& \multirow{2}{*}{\xmark} 
& \xmark 
& 58.69 & 1.70 & 906 \\
& & & \makecell{Parallel (ours)} 
& 59.67 & 20.03 & 1134 \\

\bottomrule
\end{tabular}}
\vspace{-8mm}
\end{table}

\FloatBarrier

\subsection{Ablation study}

\begin{table}[t]
\centering
\begin{minipage}[t]{0.46\linewidth}
\centering
\caption{Backbone ablation on VIGOR Same-Area. Match Lat. denotes matching-stage latency (ms), and Mem. denotes memory usage (MB), both measured under the parallel RANSAC setting.}
\label{tab:ablation_backbone} 
\scriptsize 
\setlength{\tabcolsep}{3pt}
\renewcommand{\arraystretch}{1.00} 
\resizebox{\linewidth}{!}{%
\begin{tabular}{lcccc} 
\toprule
\multirow{2}{*}{\textbf{Backbone}} & 
\multicolumn{2}{c}{\textbf{Loc. (m) $\downarrow$}} & 
\multirow{2}{*}{\makecell{\textbf{Match}\\\textbf{Lat. $\downarrow$}}} &
\multirow{2}{*}{\textbf{Mem. $\downarrow$}} \\ 
\cmidrule(lr){2-3} 
& Mean & Median & & \\ 
\midrule
RADIOv3 & \textbf{2.52} & \textbf{1.20} & \textbf{59.67} & 1134 \\ 
DINOv2 & 2.90 & 1.40 & 94.53 & \textbf{1119} \\ 
\bottomrule 
\end{tabular}%
} 
\end{minipage} 
\hfill
\begin{minipage}[t]{0.50\linewidth} 
\centering 
\caption{Ablation study on hyperparameters. The default is $\alpha=0.5$ and $N=512$.} 
\label{tab:ablation_hp} 
\scriptsize 
\setlength{\tabcolsep}{3pt} 
\renewcommand{\arraystretch}{0.95}
\resizebox{\linewidth}{!}{%
\begin{tabular}{lcc} 
\toprule 
\multirow{2}{*}{\textbf{Setting}} & 
\multicolumn{2}{c}{\textbf{Loc. (m) $\downarrow$}} \\ 
\cmidrule(l){2-3} & 
\textbf{Mean} & \textbf{Median} \\ 
\midrule
$\alpha=0$ & 2.76 & 1.23 \\ 
$\alpha=0.1$ & 2.76 & 1.24 \\ 
$\alpha=1.0$ & 3.98 & 2.50 \\ 
\cmidrule{1-3}
\textbf{Default} ($\alpha=0.5, N=512$) & \textbf{2.52} & \textbf{1.20} \\ \cmidrule{1-3} 
$N=256$ & 4.57 & 3.01 \\ 
$N=1024$ & 2.67 & 1.21 \\
\bottomrule 
\end{tabular}%
} 
\end{minipage} 
\vspace{-2mm} 
\end{table}
 
We conduct ablation studies on (i) the feature extraction backbone and (ii) hyperparameter settings. 
All experiments are conducted using the VIGOR dataset under the Same-Area.

\paragraph{\textbf{Backbone.}} We evaluate ARC-Loc's sensitivity to the feature extractor by replacing our default RADIOv3~\cite{ranzinger2024radio} with DINOv2~\cite{oquab2024dinov2}, the standard backbone for recent baselines like FG$^2$, Loc$^2$, and BevSplat.
As shown in \cref{tab:ablation_backbone}, RADIOv3 yields lower mean and median errors and faster matching-stage latency, consistent with its stable performance across context module configurations (\cref{sec:ray_gen}), likely due to its distilled training and high-resolution fine-grained cues.
ARC-Loc still remains robust and computationally efficient with DINOv2, demonstrating that our framework is not tied to a specific foundation backbone. This suggests that the proposed azimuthal ray-based formulation remains effective with different feature extractors, while the choice of backbone further affects localization accuracy and efficiency.

\paragraph{\textbf{Loss functions and hyperparameters.}} 
We conduct an ablation study on the ray-distance loss weight $\alpha$ and the number of selected matches $N$ to optimize performance under severe cross-view outliers (\cref{tab:ablation_hp}). 
While $\mathcal{L}_{\text{position}}$ alone yields a 2.76\,m mean error, the integration of $\mathcal{L}_{ray-dist}$ with our default setting ($\alpha=0.5, N=512$) is observed to improve accuracy to 2.52\,m. 
The weight $\alpha$ appears to balance the primary objective and geometric regularization; a minimal weight ($\alpha=0.1$) provides negligible gains, while an excessive weight ($\alpha=1.0$) may over-regularize the network, distracting it from the primary pose estimation objective and degrading performance to 3.98\,m. 

Similarly, the match count $N$ represents a trade-off between structural diversity and outlier noise. 
A limited count ($N=256$) appears to provide insufficient context for effective learning (4.57\,m), whereas an overly large count ($N=1024$) may introduce low-confidence matches that corrupt the training signal (2.67\,m). 
Consequently, $N=512$ is suggested as the optimal empirical balance for achieving robust localization.

\section{Conclusion}
In this paper, we presented ARC-Loc, a novel cross-view localization framework that bypasses the conventional dependencies on Bird's-Eye-View transformations and external depth foundation models. 
Inspired by the traditional human navigation technique of \textit{resection}, we reformulated ground-to-aerial correspondences as 2D line-to-point constraints and introduced the Azimuthal Ray Convergence (ARC) solver, a closed-form 2-point minimal solver that estimates the camera location directly within the native 2D image domain. 
Additionally, we proposed the ARC Loss to optimize the matching network by enforcing ray convergence without requiring dense correspondence labels or depth priors.

Through evaluations on the VIGOR and KITTI datasets, we demonstrated that ARC-Loc achieves competitive localization accuracy compared to state-of-the-art methods. By operating entirely in 2D, we showed our approach benefits from reduced inference latency and memory consumption compared to recent frameworks relying on 3D lifting or BEV projections, highlighting its practical viability for resource-constrained autonomous platforms.

\paragraph{\textbf{Limitations.}} 
\label{sec:limitations}
 While ARC-Loc offers an efficient alternative for cross-view localization, it presents two primary limitations. First, although the framework remains robust within practical orientation noise bounds (\eg, up to $\pm10^{\circ}$), position accuracy gradually degrades under larger heading uncertainties. Second, the stability of the ARC solver relies on the directional diversity of the azimuthal rays. In scenarios with a restricted field-of-view, such as the front-facing cameras in \emph{KITTI}, matched keypoints often concentrate in a narrow region, lacking angular diversity and leading to poorly constrained ray intersections (geometric degeneracy). Addressing these remains an important direction for future research.

\paragraph{\textbf{Acknowledgements.}}
This work was supported by National Research Foundation of Korea (NRF) grants funded by the Korea government (MSIT) (No. RS-2026-25486241, 50\%), by Ministry of Trade, Industry and Energy (MOTIE) and the Korea Institute for Advancement of Technology (KIAT) under the Corporate Demand-Driven Challenge and Innovative R\&D Program for Next-Generation Researchers (Grant No. RS-2026-25539646, 25\%), and by Institute of Information \& Communications Technology Planning \& Evaluation (IITP) under the Artificial Intelligence Semiconductor Support Program to Nurture the Best Talents (IITP-(2026)-RS-2023-00253914, 25\%) grant funded by the Korea government (MSIT).


\bibliographystyle{splncs04}
\bibliography{main}
\end{document}